# A Framework for Interactive Work Design based on Motion Tracking, Simulation, and Analysis

**Liang MA[1,2], Wei ZHANG[1], Huanzhang FU[1], Yang GUO[1], Damien CHABLAT[2], Fouad BENNIS[2]**

[1]*Department of Industrial Engineering, Tsinghua University, 100084, Beijing, P.R.China*

[2] *Institut de Recherche en Communications et en Cybernétique de Nantes UMR 6597 du CNRS IRCCyN, Ecole Centrale de Nantes 1, rue de la Noë - BP 92 101 - 44321 Nantes CEDEX 03, France*

**ABSTRACT**

Due to the flexibility and adaptability of human, manual handling work is still very important in industry, especially for assembly and maintenance work. Well-designed work operation can improve work efficiency and quality; enhance safety, and lower cost. Most traditional methods for work system analysis need physical mock-up and are time consuming. Digital mockup (DMU) and digital human modeling (DHM) techniques have been developed to assist ergonomic design and evaluation for a specific worker population (e.g. 95 percentile); however, the operation adaptability and adjustability for a specific individual are not considered enough. In this study, a new framework based on motion tracking technique and digital human simulation technique is proposed for motion-time analysis of manual operations. A motion tracking system is used to track a worker's operation while he/she is conducting a manual handling work. The motion data is transferred to a simulation computer for real time digital human simulation. The data is also used for motion type recognition and analysis either online or offline for objective work efficiency evaluation and subjective work task evaluation. Methods for automatic motion recognition and analysis are presented. Constraints and limitations of the proposed method are discussed.

# 1. INTRODUCTION

Automation in industry has been increased in recent decades and more and more efforts have been made to achieve efficient and flexible manufacturing. However, manual work is still very important due to the increase of customized products and human's capability of learning and adapting (Forsman, Hasson, Medbo, Asterland, & Engstorm, 2002). Accordingly, manual assembly systems will maintain their position in the future, and with increasing competition, attention to the efficiency of manual work is growing (Groover, 2007).

The aim of ergonomics is to generate working conditions that advance safety, well-being, and performance (Kroemer, Kroemer, & Kroemer-Elbert, 1994). Manual operation design and analysis is one of the key methods to improve manual work efficiency, safety, comfort, as well as job satisfaction. In conventional methods, stop-watch, record sheets, camcorders, etc., are used as observation tools on site to record the workers' operation, and then the recorded information is processed either manually or with the help of processing software to evaluate the workers' performance and workload.

Predetermined time standard (PTS) systems have been used for work measurement to estimate the time needed by qualified workers to perform a particular task at a specified level of performance. One of the most well-known PTSs is methods-time measurement (MTM), developed by Maynard and others in the late 1940s (Laring, Forsman, Kadefors, & Ortengren, 2002), in which manual operation is divided into a set of basic element motions and each element motion has a corresponding description of manual activity. This system has been used to evaluate motion time of manual operation and analyze workload of right or left hand by calculating time distribution (Chaffin, Anderson, & Martin, 1999). Maynard operation sequence technique (MOST) is a simplified system developed from MTM. MOST utilizes large blocks of fundamental motions and 16 time fragments (Dosselt, 1992). MTM has now been computerized in order to enhance efficiency of work analysis. There are several commercial software packages available to calculate expected time consumption for a planned task, such as MOST for Windows, ErgoMOST (Laring, Christmansson, Kadefors, & Ortengren, 2005).

However, in conventional methods for operation design, the design process consists of work design and work evaluation and the whole design is highly iterative and interactive, even with the computerized ergonomics software. Onsite data collection, afterward analysis, and work redesign, are time consuming and expensive to carry out. Meanwhile, the performance evaluation could be influenced by subjective factors from analysts in manual processing analysis. In addition, it is impossible to verify the design concept without an available physical prototype of the work system with these methods (Chaffin, Thompson, Nelson, Ianni, Punte, & Bowman,2001).

In order to reduce the total design cost and engineering cost and enhance the quality of design, computer-aided engineering (CAE) and digital mockup (DMU) methods have been developed to achieve rapid and virtual prototype development and testing (Chaffin, 2007). Combining DMU with digital human models (DHM), the simulated human associated with graphics could provide visualization of the work design, and shorten the design time and enhance the number and quality of design options that could be rapidly evaluated by the design analysts (Chaffin, 2002).

Virtual assembly working process has been analyzed combining with DMU and MTM methods (Bullinger, Richter, & Seidel, 2000). In their study, data glove was used as user interface to manipulate virtual objects in virtual assembly work, and MTM standard was employed to evaluate the time of assembly and cost. The gesture for the assembly operation could be automatically recognized and mapped to MTM motion. The limitation of this study was that such method was only applicable to seated operations. In a recent study, onsite video capture has been used for work measurement practice to evaluate the assembly process in long distance (Elnekave & Gilad, 2005). In this case, computer aided analysis based on MOST enhanced the motion time evaluation efficiency. However, the operation video transferred from long distance factory has to be observed and segmented by analysts with computers. Motion time analysis was also part of the output in the framework of Michigan's HUMOSIM (Chaffin, 2002), and it was also integrated into Jack (Badler, Phillips, & Webber, 1993) in Badler, Erignac, & Liu (2002) to validate the maintenance work.

Although DMU and DHM have been used to accelerate the design process, there are still several limitations. First, as pointed out by Chaffin (2007), "posture and motions of people are not well modeled in existing digital human models, and using inverse kinematics and other related robotics methods can assist in this task, but may not be sufficient, particularly if the designer does not have a profound understanding of biomechanics as well as the time to experiment with alternative postures and motion scenarios". Second, as Zha & Lim (2003) stated, the trend towards teamwork or flexible work groups further emphasizes the need for "adjustability to individual workers and the flexibility of assembly systems". Here, adjustability indicates that the working environment needs to be adjustable for different operators according to their personal characteristics, and the work design should also be changeable to adapt to the operator's physical and mental status to avoid potential risks and enhance work efficiency.

For dynamic simulation in DHM, two methods were conducted to generate the motion of the digital human: direct computer manipulation of DMU and motion tracking of a real worker's working postures while performing a task. The latter one can result in more realistic human motion than direct computer manipulation (Badler et al., 2002). Motion tracking technique is able to digitalize the real working procedure for an individual worker into motion data, which can be further processed to analyze the manual operation. It is derived from video assisted observation methods for work analysis (Elnekave & Gilad, 2005). This technique has been used in virtual assembly evaluation (Chang & Wang, 2007) with combination of rapid upper limb analysis (RULA) and digital work environment (DELMIA Software) to evaluate the potential musculoskeletal disorder risks in automobile assembly. Besides that, in virtual assembly, motion tracking has been employed to analyze the motion-time relation for evaluating work design, with emphasis on recognizing arm movements (Bullinger, et al., 2000). However, there is no application that tried to combine motion tracking, DHM and MTM for work efficiency evaluation, especially involving consideration of human body posture analysis.

In this study, a framework to evaluate manual work operations with the support of motion tracking, DHM technique, and MOST motion time method, is proposed below. Motion tracking with DHM technique can provide more realistic motion than direct computer manipulation based method. MOST method was applied to analyze manual operation sequences and evaluate the work efficiency. A prototype system was constructed to realize the concept of the framework and validate the technical feasibility.

## 2. FUNCTION STRUCTURE ANALYSIS AND FRAMEWORK

### 2.1. Architecture of the System

The architecture of the interactive work design framework is presented in Figure 1. The overall function of interactive work design is: field-independent work design, work simulation, and work inefficiency evaluation. In order to realize the overall function, there are three sub-functions to be fulfilled: motion tracking for data collection, motion data-driven operation simulation, and automatic work analysis based on the motion data.

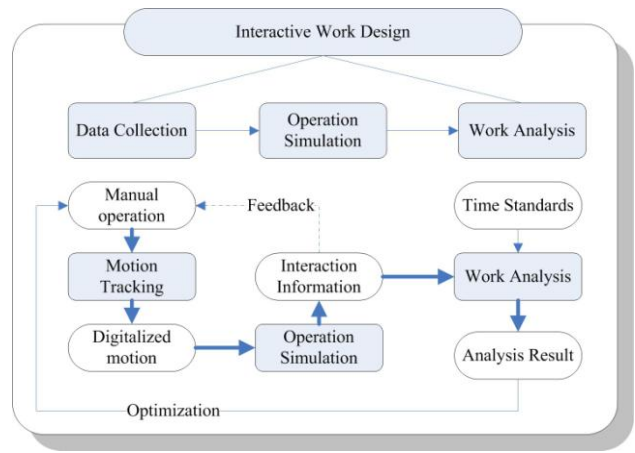

Figure 1 Framework of interactive work design system

With the support of DMU technique and DHM technique, virtual human and virtual work system can be constructed to represent the physical appearance of worker and system. Real worker can work in an immersive virtual environment, performing certain task by interacting with virtual workplace and virtual tools. In the interactive environment, motion tracking system digitalizes the worker's operation into motion data, which describes the coordinates of the worker's key joints; the data is further mapped into virtual human to manipulate the same operation in the virtual environment. Visual feedback and haptic feedback are provided to the worker in a real time manner to support the interaction. At last, all the motion data and interaction information are processed by software to assess the performance of the manual operation. In the framework, MOST analysis could be one of the analysis results.

## 2.2. Motions in MOST

In MOST standards, motions are classified into three motion groups: general move, controlled move, and tool use (Niebel & Freivalds, 1999). The general move identifies the spatial free movement of an object through the air, while the controlled move sequence describes the movement of an object when it either remains in contact with a surface or remains attached to another object during the movement. The tool use sequence is aimed at the use of common hand tools. In MOST standards, almost all the manual operation can be segmented into element actions with corresponding standard time.

In Figure 2, general motion in MOST is analyzed in details. Four parameters, action distance ($A$), body move

($B$), gain control ($G$), and placement ($P$), are used to describe a general motion. Parameter $A$ covers all the spatial movements or actions of fingers, hands and/or feet, either loaded or unloaded (Zandin, 2003). According to descriptions of $A$ moves, such as "within reach" and "1-2 steps", the horizontal trunk motion can represent most of $A$ motions; $B$ indicates body movement to achieve the action of $G$ and $P$, and "it refers to either vertical motions of the body or the action necessary to overcome an obstruction or impairment to body movement." (Zadin, 2003). $G$ refers to all the manual motions involved to obtain the complete control and the release of an object after placement. $P$ covers the actions occurring at the final stage of an object's displacement, such as alignment, position, etc.

For each parameter, different descriptions are listed to describe different movements. For example, in the column of $B$, several terms, such as "Bend", "Sit" and "Stand", are used to describe different types of body movements. One description maps to one fixed MOST index which can be used to calculate the expected time for this action using the equation $T = 10 \times index \times TMU$. TMU is time measurement unit and equals to 1/100000 h = 0.036 sec. Controlled Move and Tool Use are constructed with the same concept, but with several extra parameters to describe motions including interactions with workplace, machines, and tools.

**Basic Most Data Card**

ABG — Get
ABP — Put
A — Return

**General Move**

| Index × 10 | A<br>Action Distance | B<br>Body Motion | G<br>Gain Control | P<br>Placement | Index × 10 |
|---|---|---|---|---|---|
| 0 | ≤ 2 in. (5 cm,) | No Body Motion | No Gain Control<br>Hold | No placement<br>Hold<br>Toss | 0 |
| 1 | Within Reach | | Grasp Light Object<br>Grasp Light Objects Simo | Lay Aside<br>Loose Fit | 1 |
| 3 | 1 - 2 Steps | Sit without Adjustments<br>Stand without Adjustments<br>Bend and Arise 50% occ, | Get Non-simo<br>Get Heavy/Bulky<br>Get Blind<br>Get Obstructed<br>Free Interlocked<br>Disengage<br>Collect | Loose Fit Blind<br>Place with Adjustments<br>Place with Light Pressure<br>Place with Double Placement | 3 |
| 6 | 3 - 4 Steps | Bend and Arise | | Position with Care<br>Position with Precision<br>Position Blind<br>Position Obstructed<br>Position with Heavy Pressure<br>Position with Intermediate Moves | 6 |
| 10 | 5 - 7 Steps | Sit<br>Stand | | | 10 |
| 16 | 8 - 10 Steps | Bend and Sit<br>Climb on<br>Climb off<br>Stand and Bend<br>Through Door | | | 16 |

Figure 2 General motions in MOST standards, adapted from (Niebel & Freivalds, 1999)

## 2.3. Motion Tracking

Motion tracking module in Figure 1 is responsible to provide motion information for further simulation and analysis. Motion information consists of body motion and hand gesture information. Body motion provides general information of positions and orientation of trunk and limbs, and hand gesture information is necessary to determine some detailed manual operations, such as grasping and releasing an object.

Motion tracking techniques are based on different physical principles (Foxlin, 2002). In general, requirements for tracking technique are: tiny, self-contained, complete, accurate, fast, immune to occlusion, robust, tenacious, wireless, and cheap. But in fact, every technique today falls short on at least seven of these 10 characteristics (Welch & Foxlin, 2002). The performance requirements and application purposes are the decisive factors to select the suitable technique.

In the framework, in order to record and analyze worker's motion, coordinates of the worker's key joints should be known. In order to know the operation in detail, the movement and gesture of hands are also necessary. Regarding the tracking rate, tracking worker's operation is less challenging than tracking athlete's performance (such as running or jumping) because in general there are few quick actions in work. Thus, the tracking rate should only satisfy the visual feedback requirement, in other word, at around 25 Hz (Zhang, Lin, & Zhang, 2005). Meanwhile, no tracker is suitable for tracking full-body motion and finger motion at the same time. Therefore, hybrid tracking system is necessary in order to capture all necessary motion information.

## 2.4. Simulation System

Simulation module in Figure 1 takes motion information from the motion tracking module to construct a dynamic virtual scenario. In this scenario, a virtual human interacts with virtual working system in the same way as the real worker moves, more preferably in a real time manner. In addition, the simulation module needs to provide multi-channel feedback to the real worker in order to form a human-in-the-loop simulation (Hayward, Astley, Curz-Hernandez, Grant, & Torre, 2004). The interactive feedbacks may include visual feedback, haptic feedback, and audio feedback. General requirements for simulation are: no motion sickness feeling during simulation and real time multi-channel feedback (Wilson, 1999). Virtual human and virtual working system should be modeled to present the real worker and his operation.

## 3. MOTION ANALYSIS

Motion analysis module in Figure 1 plays the role as an ergonomist to objectively and quickly evaluate the working process. In order to accomplish this objective, motion data needs to be segmented and mapped into motion descriptions in MOST standards. In MOST standards, all the element motion has a description to represent a dynamic body change. So effort is needed to translate the tracked motion information into different corresponding motion descriptions. Then, the motion information (manual operations) can be analyzed by MOST standards.

### 3.1. Standard Static Posture Recognition and Hand Action Recognition

#### 3.1.1. Related Coordinate Systems

In order to describe human motion clearly, the following two coordinate systems are used.

- World coordinate system (WCS)

In the tracking system, all three-dimensional (3D) positions are calculated relative to a world reference Cartesian coordinate system. In this coordinate system, the positive $Z$ direction is vertically up. The positive $X$ and $Y$ directions can be determined by the user when setting up the tracking system.

- Body-based trunk coordinate system (BTCS)

This coordinate system is fixed on the trunk. The line intersected by the sagittal plane and the coronal plane is a vertical line. The intersecting point between this line and the line connecting the two shoulders is selected as the BTCS origin. The positive $Z$ axis is pointing from the origin to the pelvis along the intersection line; the positive $X$ axis is pointing from right shoulder to left shoulder in the coronal plane, and the positive $Y$ axis is pointing to the front of the trunk (Figure 3).

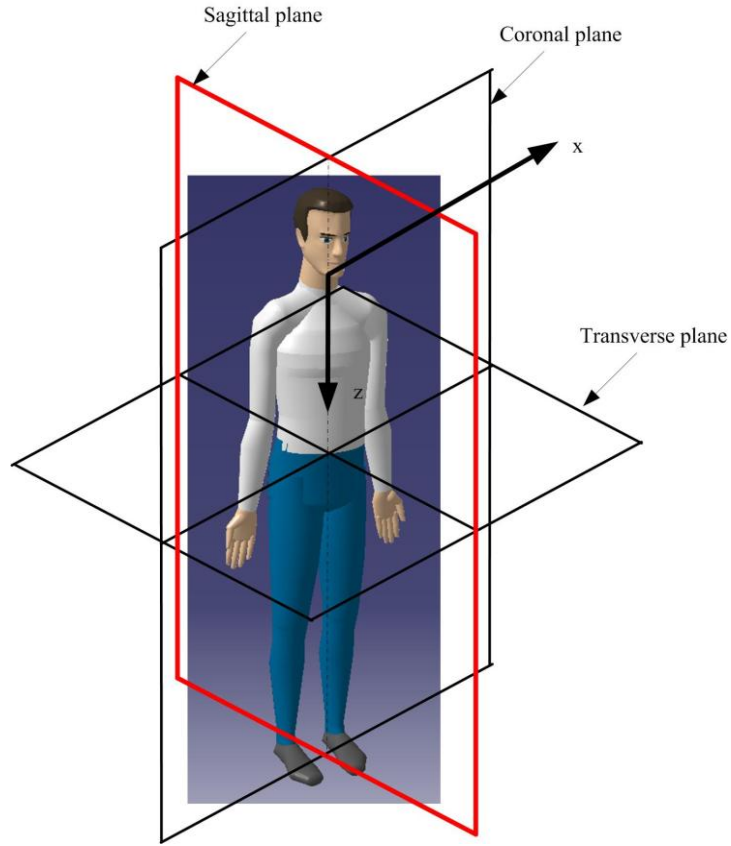

Figure 3 Body-based trunk coordinate system

The position of the trunk relative to WCS can be determined by calculating the displacement of the BTCS origin. The orientation of the trunk relative to WCS can also be calculated. Each limb can be represented by a vector pointing from proximal point to distal point. With this vector information, the position and orientation of each limb relative to BTCS can be determined. In addition, several important angles can be calculated with the orientation and position information, such as the angle between upper leg and lower leg, and the angle between upper arm and forearm. These parameters are listed in Table 1.

Table 1 Parameters to describe the static posture in body-based human coordinate system

|  | Parameters |
|---|---|
| BTCS position change | Origin of current BTCS in the original BTCS ($x_B$, $y_B$, $z_B$) |
| BTCS orientation change | ($\alpha_B$, $\beta_B$, $\gamma_B$): angles between +$X$ axis, +$Y$ axis, +$Z$ axis of current BTCS and the corresponding axes of the original BTCS. |
| Orientations of upper arm (UA), forearm (FA), upper leg (UL), lower leg (LL) | ($\alpha_{UA}$, $\beta_{UA}$, $\gamma_{UA}$): angles between the vector of upper arm and the XYZ axes of the current BTCS. |
|  | ($\alpha_{FA}$, $\beta_{FA}$, $\gamma_{FA}$): angles between the vector of forearm and the XYZ axes of the current BTCS. |
|  | ($\alpha_{UL}$, $\beta_{UL}$, $\gamma_{UL}$): angles between the vector of upper leg and the XYZ axes of the current BTCS. |
|  | ($\alpha_{LL}$, $\beta_{LL}$, $\gamma_{LL}$): angles between the vector of lower leg and the XYZ axes of the current BTCS. |
| Arm angle | $\theta_A$: angle between vector of upper arm and vector of forearm |
| Leg angle | $\theta_L$: angle between vector of upper leg and vector of lower leg |

### 3.1.2. Recognition of Standard Static Postures and Hand Actions

In the proposed framework, motion information is recorded frame by frame. In each frame, the motion information can only describe a static posture. In order to recognize a dynamic working motion, as defined in MOST, only one static posture in one frame is not enough. But rather, a series of static postures are needed to recognize a dynamic motion.

In this study, recognition of dynamic working motions is accomplished in two steps. In step 1, according to each frame of motion information, a standard static posture is defined; in step 2, finding out the frame where standard static posture changes (e.g. from sitting to standing). Whenever there is a standard static posture change, it is a start or end of a standard motion.

Definition of the standard static postures consists of two parts: natural-language description (for human) and technical parameters (for computer). For example, to define "sitting" posture, the natural language description is "sitting", and the parameters include the position and orientation of trunk and limbs. In Table 2, definition of "sitting" posture is given. It begins and ends with syntax "Rules" and "\Rules", respectively. The second line tells that the parameter in MOST standards is "B" and the natural language description is "sitting". The third line tells that there are three keywords ("Trunk", "Left Leg", and "Right Leg"). Each keyword has a defined parameter (such as "Z Axis" or "Relative Angle") with its corresponding mean value and allowable variance. Other static postures are defined in similar ways except that their corresponding keywords, parameters, and values are different. The following pseudo codes define "sitting" posture using the information in Table 2.

```
IF      Trunk.Z_Axis >-5 AND Trunk.Z_Axis<+5
    AND        LeftLeg.RelativeAgnle>(90-20)          AND
LeftLeg.RelativeAgnle<(90+20)
    AND        RightLeg.RelativeAgnle>(90-20)         AND
RightLeg.RelativeAgnle<(90+20)
THEN    CurrentPosture ASSIGNED AS sitting
ENDIF
```

Table 2 Standard static posture definition of "sitting"

| Definition of "sitting" static posture | | | | |
|---|---|---|---|---|
| Start syntax | Rules | | | |
| Motion group | B | sitting | | |
| Involved body segments and | Trunk | Z Axis "(α)" | 0 "(mean value)" | 5 "(variance)" |
| | Left Leg | Relative Angle (θ) | 90 | 20 |
| | Right Leg | Relative Angle (θ) | 90 | 20 |
| | ⋮ | ⋮ | ⋮ | ⋮ |
| End syntax | \Rules | | | |

For hand gesture, different close or open degrees of fingers represent different gestures. In this framework, the gesture information, obtained from data glove, is not important unless there is an interaction between the worker and the virtual working system. Specific action of hand can be determined using the gesture data and collision information. For example, "touching a button" includes pointing gesture of index finger (showing the intention) and collision between the index finger and a virtual button. Otherwise, the gesture is meaningless for operation analysis. Collision information is obtained by analyzing the collision between human and objects in the working system, or between two objects in the system. In general, the collision can be further classified into two types: type I (collision between virtual human and a virtual object) and type II (collision between two virtual objects). For type I, each virtual object colliding with virtual human has its corresponding function. According to its function, the corresponding collision information can be determined. For example, a collision with a switch button is defined as "touching button". As shown in Table 3, "touching button" is defined as: hand "touching" a button (virtual object) with a pointing index gesture. The identification (ID) of the virtual object can be predefined when constructing the virtual working system. "Positive" in Table 3 means from "no contact" between the fingertip and the virtual object to "contact". For type II, the collision between two objects can occur in the "tool use" motions. For example, "fastening" and "cutting" operation in MOST standards indicates collision between two objects. In these motions, human operator has to be involved, but the motion can only be detected unless there is collision between tools and objects.

Table 3 Hand action definition of "touching button"

| Start syntax | Rules | | | |
|---|---|---|---|---|
| Motion group | G | Touching button | | |
| Interaction Information | Virtual Object | Virtual ID | | |
| | Interaction Part | Hand | Gesture | Pointing Index |
| | Trigger | Positive | | |
| End syntax | \Rules | | | |

### 3.2. Motion Recognition

After standard static posture recognition and hand action recognition, the overall working process is segmented automatically by the key frames (where there is posture change). The transaction between two successive key frames represents a dynamic motion. For example, the process from "sitting" static posture to "standing" static posture is the motion "stand" in MOST. Based on this concept, a motion matrix is constructed for "$B$" motions to describe the corresponding motion from "Static Posture $i$ to" "Static Posture $j$". The content in the cell of the matrix is also a natural language description. A simplified motion matrix is shown in Table 4. The first column lists the static postures of the previous key frame, and the first row lists the static postures of the current key frame. The intersection value represents the dynamic motion process from the static posture of the previous key frame to the new static posture of the current key frame. The following pseudo codes define "stand" motion in MOST standard.

```
IF      start posture EQUALS TO sitting AND end posture EQUALS TO
standing
THEN    motion ASSIGNED AS stand
ENDIF
```

Table 4 Dynamic motion definition matrix based on standard static postures

|  | "Standing" | "Sitting" | "Bending" |
|---|---|---|---|
| From "standing" to | No motion | Sit | Bend |
| From "sitting" to | Stand | No motion | Bend |
| From "bending" to | Arise | Arise | No motion |

For "$A$" motions in MOST, it is easy to decide the action distance by calculating the displacement of BTCS relative to WCS. Motion "$A$" is the horizontal displacement of trunk between two key frames. When key frames are determined automatically, the action distance is calculated at the same time by software. The only thing that needs to consider is whether it is necessary to calculate the action distance. The "action distance" of human body must have its corresponding aim for the movement. For example, "move 4 steps (approximate 3 m)" to take a wrench tool. Otherwise, the movement without an aim is time loss in the working process.

For hand motion, gesture and collision information are both used to determine the motion type. For example, the period between "touching button" and "releasing button" is a processing time for button control. And the period in which two virtual objects are engaged can be recognized as "tool use" time. "G" and "P" motions can also be recognized with collision information.

### 3.3. Motion-Time Analysis

Each MOST description has a corresponding index $p$. The standard time required for a motion is $0.036p$ [s]. The actual time consumed by the real worker motion can be calculated from the start key frame number $n_s$ and the end key frame number $n_e$ as $(n_e-n_s)T$ [s], where $T$ is the interval of two successive frames. Therefore, the efficiency of each motion can be evaluated by comparing the actual consumed time with the standard required time. The overall procedure is shown in Figure 4.

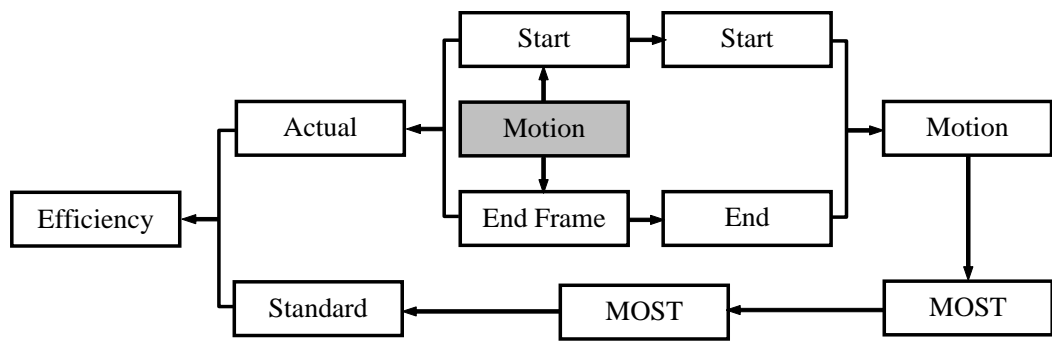

Figure 4 Flow chart from motion data to efficiency evaluation

## 4. PROTOTYPE SYSTEM REALIZATION

### 4.1. Hardware and software system

The above two sections give general requirement, principles, and procedures of the proposed framework,

which is independent of the hardware and software system. This section will demonstrate a developed prototype system with specific hardware and software configurations in order to realize the proposed system and validate the technical feasibility.

The hardware structure is shown in Figure 5. In this system, optical motion tracking system was employed to capture human motion, while 5DT data glove (produced by Fifth Dimension Technologies) was used to track hand motion. The tracked data was transferred via network to simulation computer to realize real time simulation. Audio and visual feedback was provided to the worker by head mounted display (HMD). Haptic feedback was realized through clothes-embedded micro vibration motors. Projection-based wide screen display was also used for supervisor or other third party to evaluate the system.

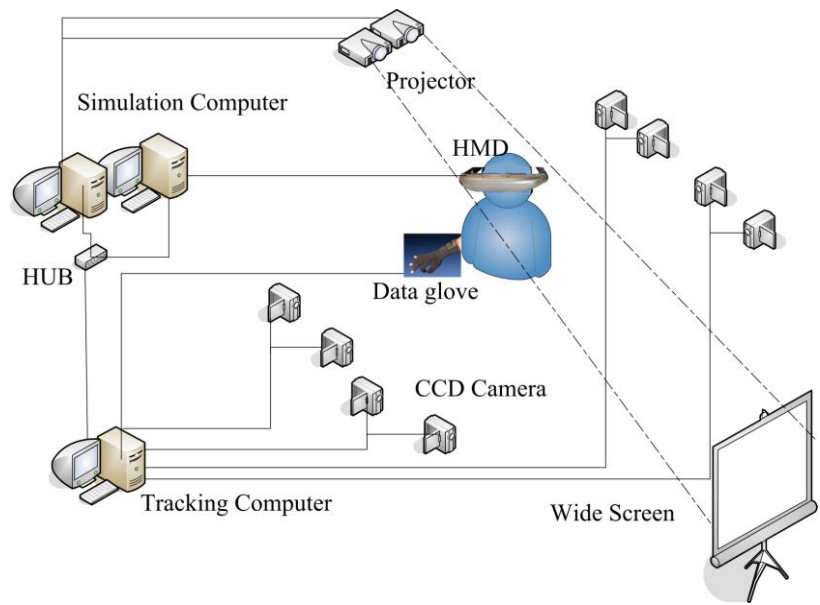

Figure 5 Hardware scheme of the framework prototype system

The optical motion tracking system was equipped with eight CCD cameras around the work space in order to minimize line of sight problem. Optical motion tracking is a technique that uses stereoscopic vision principle to calculate markers' 3-dimensional coordinates. Usually a marker to be tracked needs to be "seen" by at least two cameras. The captured image is processed by computer to identify the marker position in each image (2-dimensional pixel coordinates). By comparing the pixel coordinates with prior calibration information, the 3D coordinates can be calculated in real time (frame by frame). The time resolution of MOST is 0.36 [s] ($10 \times$TMU), so the identifiable time segment should be smaller than this resolution requirement. In this prototype system, each CCD worked at 25 Hz, so the time period between each frame (totally 25*8 = 200 frames per second with 8 CCDs) was 1/200 = 0.005 [s], and for each CCD, the time period was 1/25 = 0.04 [s], both of which satisfied the time resolution requirement. However, with the frame rate of 25 Hz, the capture system was not suitable for fast motion tracking, such as running and emergency operation. Fortunately there are very rare fast motions in regular manual handling operations.

Spatial accuracy of motion tracking system affects motion recognition reliability, especially for small range movement. The accuracy of the optical capture system depends on its hardware setup and the algorithms in the capture module. The absolute accuracy of our prototype, within 5 cm depending on the location of the target in the WCS, was relatively low. However, the repeatability was much higher, around 2-3 mm. This ensured the analysis of human motion was reasonably reliable. Repeatability means the error between different times of measurement. As long as the measurement is reasonably consistent among different times of measurement (repeatability), motion recognition can be carried out even with larger error of absolute accuracy from actual coordinate values. Usually, the most important issue in optical motion tracking is to avoid occlusion of optical markers (light of sight problem) and confusion of markers (such as when moving the left hand near the right hand, or somewhere else of the body key joints). Our hardware setup and algorithms could usually track a worker's motion (with 13 markers on the key joints) for around 3-5 minutes stably. This had made it possible to track many working operations within such a time period. Figure 6 shows two real worker postures and their corresponding simulated postures based on the tracking data.

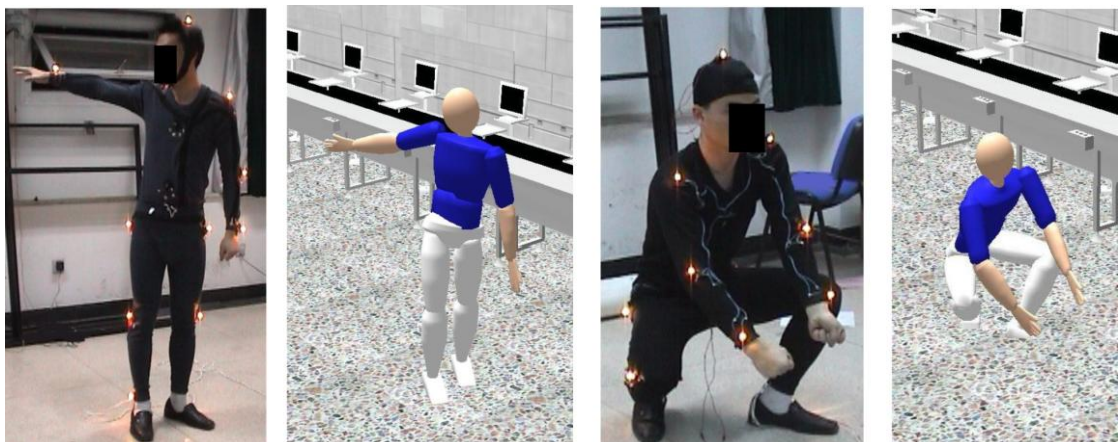

Figure 6 Real postures and the corresponding simulated postures

Data Glove 5 Ultra, with five sensors, produced by Fifth Dimension Technologies, was employed in the motion capture system to record hand motion. Compared to body motion tracking, our hand gesture tracking was much more accurate and had less trouble. Each finger had one corresponding fiber optics based sensor, which had a resolution of 10bit (1/1024). In principle, recognizable gestures are determined by the number of sensors in data gloves. With five sensors, only 16 gestures could be determined considering the open and close status of the sensors. This limitation resulted in some difficulties in determining the detailed interaction between human and the virtual objects.

Three software tools were developed to perform the functions of three defined modules in Figure 1: motion tracking, motion simulation, and motion analysis. Motion tracking software was responsible to process the captured images from each CCD, calculating the motion data based on prior calibration data and current marker position in each image, transferring motion data to simulation computer, and saving the motion data for further analysis. The simulation software was responsible to create the immersive virtual environment by providing visual and audio feedbacks and to enable the interaction between human and virtual objects via haptic feedbacks. The motion analysis software was responsible to retrieve the saved motion data and interaction information and to recognize the standard static postures, dynamic motions, and evaluate the efficiency. The interfaces of the three software tools, all developed in Visual C++ under Windows XP, are shown in Figure 7. MultiGen Creator and Vega software package were used to create and simulate the virtual working environment.

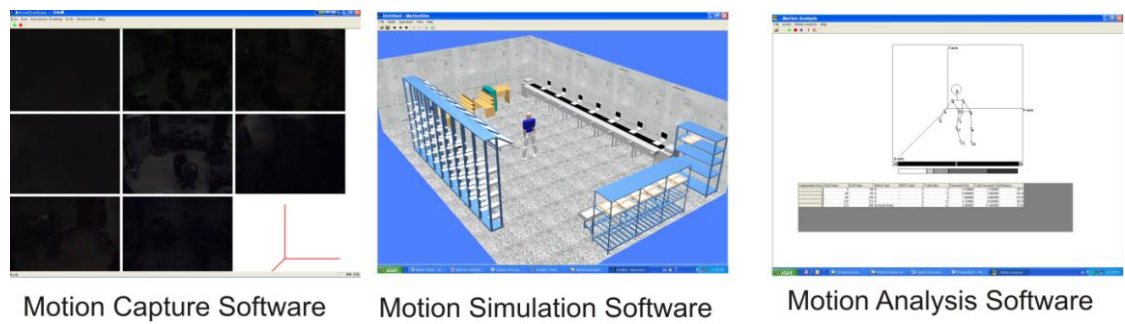

Figure 7 Three software tools developed for the prototype system

### 4.2. Typical Work Tasks

Several typical work tasks were taken to validate the technical feasibility of the proposed framework. Lifting was taken to test the general motions, and pushing button was taken to test controlled moves. In general, a lifting operation has four subsequences: moving to the work piece [*A*], bending down [*B*], grasping the work piece [*G*], arising up [*B*], and releasing the workpiece [*P*]. For pushing button, it includes: moving to button area [*A*], touching the button [*G*], processing, and releasing the button [*P*], and back [*A*].

Lifting operation was taken to validate the technical feasibility of the motion analysis tool. The purpose of this validation was to check whether the developed prototype system could function for a lifting task. If the system could successfully track an operator's moves and the moves could be correctly extracted from the tracking data, then we will think it technically feasible. Usability of the system (how usable for different people) is not a concern at this time. As stated before, a lifting operation composes of a sequence ABGBP. The lifting operation was performed by a subject in a virtual working environment in laboratory condition with no *A* motion (no walking). The motion data was captured by the capture system. Once the operation was completed, the tracking data and interaction information were analyzed by the analysis software tool. In Figure 8, analysis result of the lifting operation is presented in the form of a table. The different stages of the lifting operation are graphically shown above the table which reflects the real motion (from 1 to 6). The change of the trunk orientation was recognized as 6 standard static postures. From the result, it could be found that the whole working process was segmented into 5 sections (dynamic motions). They are B, BG, B, BP, and No Motion. The first B motion represents the half bend of trunk, and BG means fully bending body while grasping the work piece. Raising up and releasing the object were recognized as B and BP. Finally, after completing the job, the subject remained standing for a moment. This period was recognized as "No motion" in the software, because it had no contribution to the lifting operation. The efficiency of the first segment and the last segment was very low, much less than 100%, because at the beginning and the end the subject stood for a little before and after the lifting operation. The efficiency of the other segments was a little low because the subject had some difficulty in interacting with virtual object.

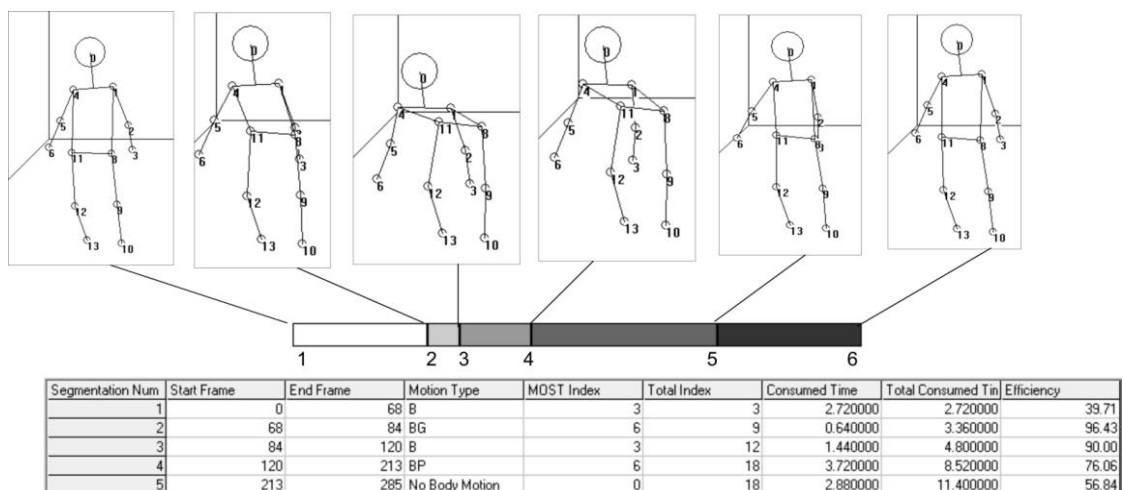

Figure 8 Analysis results for a lifting job

The moves involving "pushing button" was also successfully tracked and analyzed, as illustrated in Wang,

Zhang, Bennis, & Chablat (2006). Figure 9 shows two simulated motions driven by the tracked motions of a real worker to perform some moves involving collision information (moving an object from right to left, and open-close a door, which are also like pushing button because collision detection and response are required).

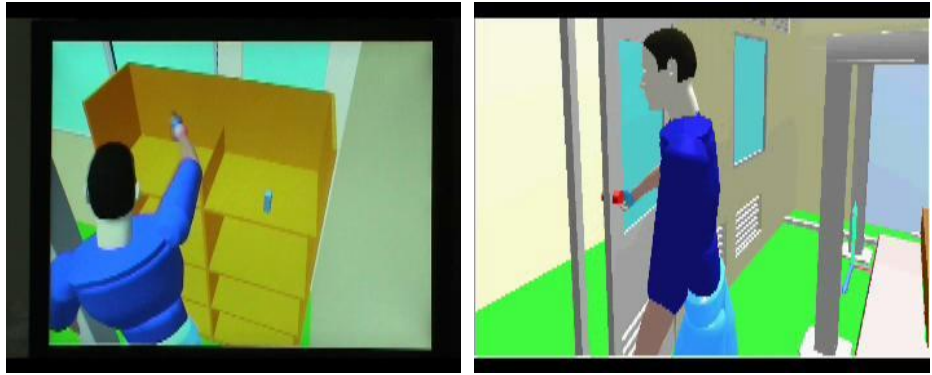

Figure 9 Virtual human moving an object

### 4.3. Limitation of Recognizable Motions

Not all the motion descriptions in MOST standards can be recognized automatically by the motion analysis software, due to insufficient information (either physical or mental) that was available in the prototype system. In the following, limitations of recognizable motions will be discussed. With the current prototype system setup, some MOST motions were recognizable with no problems, and some were recognizable with limitations, and some were even unrecognizable.

- Recognizable motions

In general, action distance (*A*) can be recognized with no problem. It indicates the displacement of the human body. The current tracking system had the trunk motion data and had no difficulty to calculate the horizontal displacement of trunk. Usually for any meaningful task, "*A*" motion accompanies other principal motions to achieve a certain goal. So "*A*" motion is not an independent motion, but rather an assisting motion to move the human body to the operation zone of the principal motion. Accordingly, "*A*" is calculated after the principal motion is segmented.

"B" motion concerns the postural change of human body. It can be further divided into two groups: one group without interactions with working system and one with interactions. In the B column of Figure 2, "Bend and Arise 50%", "Bend and Arise", "Sit", Stand", "Bend and Sit", and "Bend and Stand" belong to the first group and can be recognized automatically with no problem.

- Recognizable motions with limitations

Some other B motions in the column of Figure 2 belong to the second group, such as "climb on", "climb off", and "through door". These motions can not be recognized directly from the motion data, but rather need additional interaction information. So these kinds of motions were recognizable with some limitations using the prototype system.

For controlled move and tool use, with combination of interaction information, these motions are recognizable. However, in the current prototype system, only some interactions were realized due to the technical limitation of the current data glove. These interactions are control moves, such as pushing button, grasping object (Wang et al., 2006), and tool use moves, such as fastening or loosening a screw

- Unrecognizable motions

In MOST, there are several tool use motions labeled as "thinking", "reading", and "inspecting". These motions describe mental efforts in manual operations. During these operations, there is no physical interaction between human and working system. Therefore, these "motions" can not be recognized automatically.

Besides the above mentioned motions describing mental efforts, several motions in the "general move" are also difficult to recognize even with interaction information. For example, "sit without adjustments" and "stand with adjustments" can only be recognized as "sit" and "stand", because "adjustment" is too general to describe for computer programs.

For all the unrecognizable motions, manual modification is needed. In the prototype system, key frames can be added or removed via the software interface to re-segment the whole process and assign a motion type.

### 4.4. Other Limitations

System limitations are mainly from tracking techniques and simulation techniques. The work analysis is carried out based on the data from motion tracking and simulation data. Therefore, the fidelity of the simulation system may influence the worker's performance because of additional effort required to judge distance, especially depth cue. The accuracy and repeatability of tracking system has great influence on real worker's motions and efforts. Lower accuracy and repeatability would make the worker's operation difficult and need more effort. In addition, there are countless kinds of interactions between human and machine, and it is impossible to realize all the possible interactions in a virtual working environment. As a result, only interactions involved in the manual operations need to be considered in the framework. As mentioned above, if interaction information is accurate and sufficient, most motions in MOST standards can be correctly recognized, especially for controlled move and tool use motions. However, the fidelity of virtual reality system depends on many factors, and it is effortful to provide a high-end immersive setting.

In MOST standards, only natural language descriptions are used to describe the element motions. As discussed before, a natural description has too many posture possibilities. So in the future, to make the proposed framework feasible for more conditions, it is necessary to develop a good way to parameterize the descriptions in MOST, including trying to incorporate more interaction information.

## 5. CONCLUSION

The framework proposed a digital way of analysis and optimization of working postures, worker subjective evaluation of system design, and evaluation of workload. Motion tracking technique is applied to digitize human's operation motions, which enables human-in-the-loop simulation in real time and further enables semi-automatic motion analysis. System functions and their requirement are discussed. A prototype system has been constructed to realize the framework and validate the technical feasibility. The results show that many general motions in MOST could be recognized while some special motions, which either need extensive interaction information or are too difficult to parameterize, could not be recognized by the prototype system. Special effort is needed in the future to develop better models to parameterize the MOST motion descriptions.

**ACKNOWLEDGEMENTS**

The authors would like to acknowledge the financial support from NSFC (National Natural Science Foundation of China, under grant number 50205014) and Mitsubishi Heavy Industry to realize the prototype system. Support from EADS and the Region des Pays de la Loire (in France) promoted the application of the prototype system in ergonomics study through collaboration between the Ecole Centrale de Nantes and Tsinghua

University.

**REFERENCES**

Badler, N. I., Erignac, C. A., & Liu, Y. (2002). Virtual humans for validating maintenance procedures. Communication of the ACM, 45(7), 57-63.

Badler, N. I., Phillips, C. B., & Webber, B. L. (1993). Simulating humans. New York: Oxford University Press, Inc.

Bullinger, H. J., Richter, M., & Seidel, K.-A. (2000). Virtual assembly planning. Human Factors and Ergonomics in Manufacturing, 10(3), 331-341.

Chaffin, D. B., Thompson, D., Nelson, C., Ianni, J., Punte, P. & Bowman, D. (2001). Digital human modeling for vehicle and workplace design. Society of Automotive Engineer.

Chaffin, D. B. (2002). On simulating human reach motions for ergonomics analyses. Human Factors and Ergonomics in Manufacturing, 12(3), 235-247.

Chaffin, D.B. (2007). Human motion simulation for vehicle and workplace design. Human Factors and Ergonomics in Manufacturing, 17(5), 475-484.

Chaffin, D. B., Anderson, G. B. J., & Martin, B. J. (1999). Occupational biomechanics (third ed.). Wiley-Interscience.

Chang, S.-W., & Wang, M.-J. J. (2007). Digital human modeling and workplace evaluation: using an automobile assembly task as an example. Human Factors and Ergonomics in Manufacturing, 17(5), 445-455.

Dosselt, R. (1992). Computer application of a natural-language predetermined motion time system. Computers & Industrial Engineering, 23(1), 319-322.

Elnekave, M., & Gilad, I. (2005). Long distance performance time inputs for global work measurement practice. The International Journal of Industrial Engineering, 12(4), 314-321.

Forsman, M., Hasson, G.-A., Medbo, L., Asterland, P., & Engstorm, T. (2002). A method for evaluation of manual work using synchronized video recordings and physiological measurements. Applied Ergonomics, 33(6), 533-540.

Foxlin, E. (2002). Motion tracking requirements and technologies. In Stanney, K. (Ed.), Handbook of Virtual Environment Technology (pp. 163-210). New Jersey: Mahwah.

Groover, M. P. (2007). Automation, production systems, and computer-integrated manufacturing (third ed.). Prentice Hall.

Hayward, V., Astley, O. R., Curz-Hernandez, M., Grant, D., & Torre, G. R.-D.-L. (2004). Haptic interfaces and devices. Sensor Review, 24(1), 16-29.

Kroemer, K., Kroemer, H., & Kroemer-Elbert, K. (1994). Ergonomics: how to design for ease and efficiency. Englewood Cliffs, NJ: Prentice-Hall.

Laring, J., Christmansson, M., Kadefors, R., & Ortengren, R. (2005). ErgoSAM: a preproduction risk identification tool. Human Factors and Ergonomics in Manufacturing, 15(3), 309-325.

Laring, J., Forsman, M., Kadefors, R., & Ortengren, R. (2002). MTM-based ergonomics workload analysis. International Journal of Industrial Ergonomics, 30(3), 135-148.

Niebel, B. W., & Freivalds, A. (1999). Predetermined time systems. In Niebel, B. W., & Freivalds, A. (Eds.), Methods, Standards and Work Design (pp. 492-498). McGraw-Hill.

Wang, Y., Zhang, W., Bennis, F., & Chablat, D. (2006). An integrated simulation system for human factors study. Industrial Engineering Annual Conference. Orlando, Florida, USA.

Welch, G., & Foxlin, E. (2002). Motion tracking: no silver bullet, but a respectable arsenal. IEEE Computer Graphics and Application, 22(6), 24-38.

Wilson, J. R. (1999). Virtual Environments Applications and Applied Ergonomics. Applied Ergonomics, 30(1), 3-9.

Zandin, K. B. (2003). MOST work measurement systems. CRC Press.

Zha, X. F. & Lim, S. Y. E. (2003). Intelligent design and planning of manual assembly workstations: A neuro-fuzzy approach. Computers & Industrial Engineering, 44(4), 611-632.

Zhang, W., Lin, H., & Zhang, W. H. (2005). From virtuality to reality: individualized freeform model design and rapid manufacturing. Human Factors and Ergonomics in Manufacturing, 15(4), 445-459.